\pdfoutput=1

\documentclass[11pt]{article}

\usepackage{acl}

\usepackage{times}
\usepackage{latexsym}

\usepackage[T1]{fontenc}

\usepackage[utf8]{inputenc}

\usepackage{microtype}

\usepackage{graphicx}
\usepackage{bm}
\usepackage{amssymb}
\usepackage{url}

\usepackage[T1]{fontenc}

\usepackage[utf8]{inputenc}

\usepackage{microtype}

\newcommand{\baseline}{\textsc{baseline}}
\newcommand{\bagging}{\textsc{bagging}}
\newcommand{\baselt}{\textsc{base-lt}}
\newcommand{\activelt}{\textsc{active-lt}}
\newcommand{\randomlt}{\textsc{random-lt}}
\newcommand{\anylt}{\textsc{*-lt}}

\newcommand{\beatbase}[1]{\emph{#1}}
\newcommand{\beatblt}[1]{\emph{\textbf{{#1}}}}
\newcommand{\bestorpar}[1]{\underline{#1}}

\title{Diverse Lottery Tickets Boost Ensemble from a Single Pretrained Model}

\author{Sosuke Kobayashi\textsuperscript{1,2} \quad Shun Kiyono\textsuperscript{3,1} \quad Jun Suzuki\textsuperscript{1,3} \quad Kentaro Inui\textsuperscript{1,3} \\
  Tohoku University\textsuperscript{1} \quad
  Preferred Networks, Inc.\textsuperscript{2} \quad
  RIKEN\textsuperscript{3} \\
  \texttt{sosk@preferred.jp} \,\,
  \texttt{shun.kiyono@riken.jp} \\
  \texttt{jun.suzuki@tohoku.ac.jp} \,\, \texttt{inui@tohoku.ac.jp}  }

\begin{document}
\maketitle
\begin{abstract}
Ensembling is a popular method used to improve performance as a last resort.
However, ensembling multiple models finetuned from a single pretrained model has been not very effective; this could be due to the lack of diversity among ensemble members.
This paper proposes \textit{Multi-Ticket Ensemble}, which finetunes different subnetworks of a single pretrained model and ensembles them.
We empirically demonstrated that winning-ticket subnetworks produced more diverse predictions than dense networks, and their ensemble outperformed the standard ensemble on some tasks.

\end{abstract}

\section{Introduction}
\emph{Ensembling}~\cite{levin1989ensemble,domingos1997baysian} has long been an easy and effective approach to improve model performance by averaging the outputs of multiple comparable but independent models.
\citet{allenzhu2020understanding} explain that different models obtain different views for judgments, and the ensemble uses complementary views to make more robust decisions.
A good ensemble requires diverse member models.
However, how to encourage diversity without sacrificing the accuracy of each model is non-trivial~\cite{liu1999negcorr,kirillov2016tutorial,rame2021dice}.

\begin{figure}[t]
  \begin{center}
  \includegraphics[width=7.5cm]{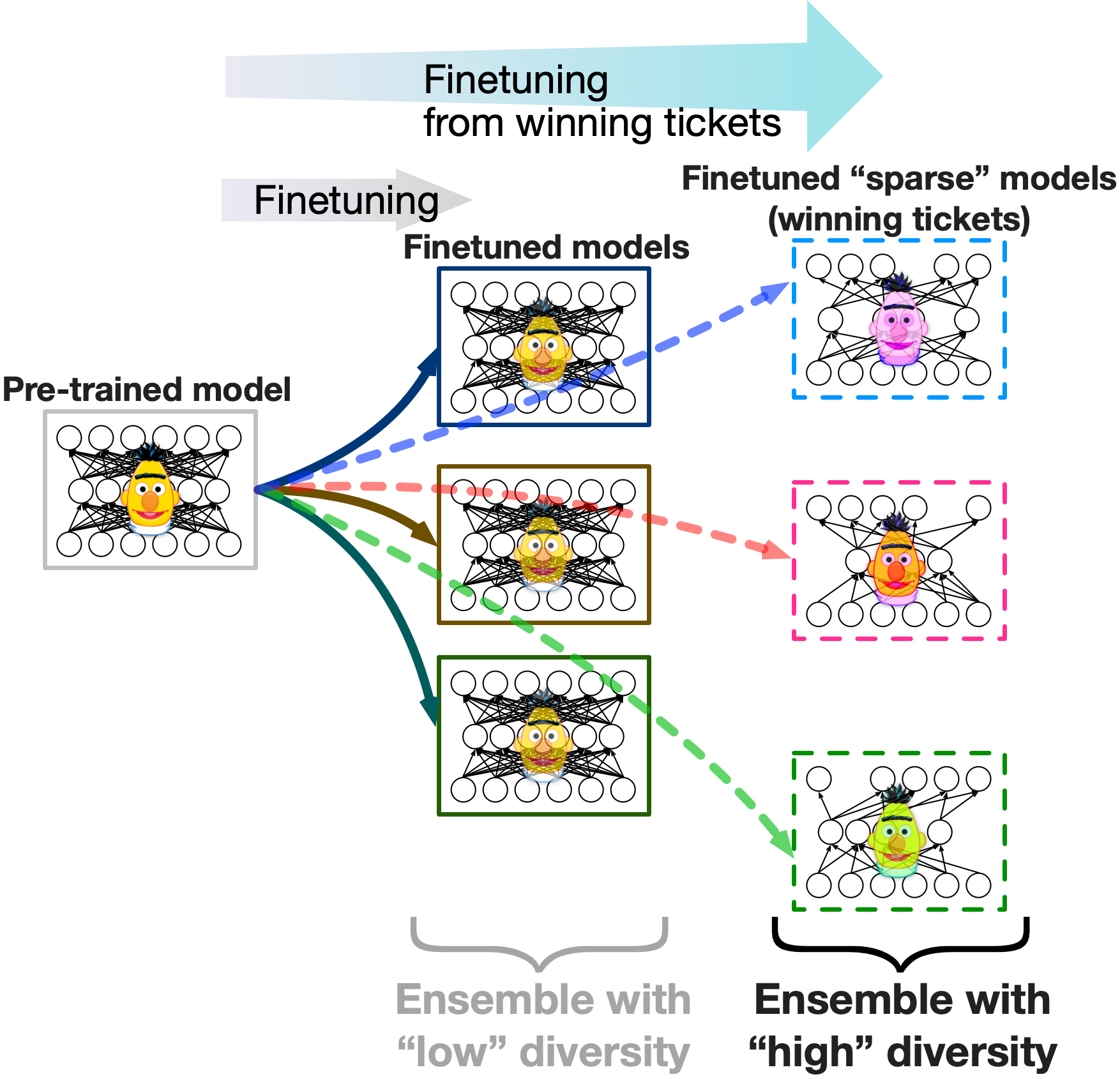}
    \caption{When finetuning from a single pretrained model (left), the models are less diverse (center). If we finetune different sparse subnetworks, they become more diverse and make the ensemble effective (right).}
    \label{fig:overview}
  \end{center}
\end{figure}

The \emph{pretrain-then-finetune} paradigm has become another best practice for achieving state-of-the-art performance on NLP tasks~\cite{devlin2019bert}.
The cost of large-scale pretraining, however, is enormously high~\cite{sharir2020cost};
This often makes it difficult to independently pretrain multiple models.
Therefore, most researchers and practitioners only use a \emph{single} pretrained model, which is distributed by resource-rich organizations.

This situation brings up a novel question to ensemble learning: Can we make an effective ensemble from only \emph{a single pre-trained model}?
Although ensembles can be combined with the pretrain-then-finetune paradigm, an ensemble of models finetuned from \emph{a single pretrained model} is much less effective than that using \emph{different pretrained models from scratch} in many tasks~\cite{raffel2020limittransfer}.
Naïve ensemble offers limited improvements, possibly due to the lack of diversity of finetuning from the same initial parameters.

In this paper, we propose a simple yet effective method called \emph{Multi-Ticket Ensemble}, ensembling finetuned \emph{winning-ticket subnetworks}~\cite{frankle2018the} in a single pretrained model.
We empirically demonstrate that pruning a single pretrained model can make diverse models, and their ensemble can outperform the naïve dense ensemble if winning-ticket subnetworks are found.

\section{Diversity in a Single Pretrained Model}

In this paper, we discuss the most standard way of ensemble, which averages the outputs of multiple neural networks; each has the same architecture but different parameters.
That is, let $f(\bm{x};\bm{\theta})$ be the output of a model with the parameter vector $\bm{\theta}$ given the input $\bm{x}$, the output of an ensemble is $f_{\mathcal{M}}(\bm{x}) = \sum_{\bm{\theta} \in \mathcal{M}} f(\bm{x};\bm{\theta}) / |\mathcal{M}|$, where $\mathcal{M} = \{\bm{\theta}_1, ..., \bm{\theta}_{|\mathcal{M}|}\}$ is the member parameters.

\subsection{Diversity from Finetuning}\label{sec:div_finetuning}

As discussed, when constructing an ensemble $f_{\mathcal{M}}$ by finetuning from a single pretrained model multiple times with different random seeds $\{s_1, ..., s_{|\mathcal{M}|}\}$, the boost in performance tends to be only marginal.
In the case of BERT~\cite{devlin2019bert} and its variants, three sources of diversities can be considered: random initialization of the task-specific layer, dataset shuffling for stochastic gradient descent (SGD), and dropout.
However, empirically, such finetuned parameters tend not to be largely different from the initial parameters, and they do not lead to diverse models~\cite{radiya2020fine}.
Of course, if one adds significant noise to the parameters, it leads to diversity; however, it would also hurt accuracy. %

\subsection{Diversity from Pruning}\label{sec:div_pruning}

To make models ensuring both accuracy and diversity, we focus on subnetworks in the pretrained model.
Different subnetworks employ different subspaces of the pre-trained knowledge~\cite{radiya2020fine,zhao2020masking,cao2021prob}; this would help the subnetworks to acquire different views, which can be a source of desired diversity\footnote{
Some concurrent and recent studies also investigate subnetworks for effective ensemble~\cite{durasov2021masksembles, havasi2021training} for training-from-scratch settings of image recognition.
}.
Also, in terms of accuracy, recent studies on the \emph{lottery ticket hypothesis}~\cite{frankle2018the} suggest that a dense network at initialization contains a subnetwork, called the \emph{winning ticket}, whose accuracy becomes comparable to that of the dense one after the same training.
Interestingly, the pretrained models including BERT also has a winning ticket for finetuning on downstream tasks~\cite{chen2020bertlth,chen2021lthsupselfcv}.
Thus, if we can find \emph{diverse winning tickets}, they can be good ensemble members with the two desirable properties: diversity and accuracy.

\section{Subnetwork Exploration}\label{sec:pruning}

We propose a simple yet effective method, \emph{multi-ticket ensemble}, which finetunes different subnetworks instead of dense networks.
Because it could be a key how to find subnetworks, we explore three variants based on iterative magnitude pruning.

\subsection{Iterative Magnitude Pruning}\label{sec:imp}

\begin{figure}[t]
  \centering 
  \includegraphics[width=7.3cm]{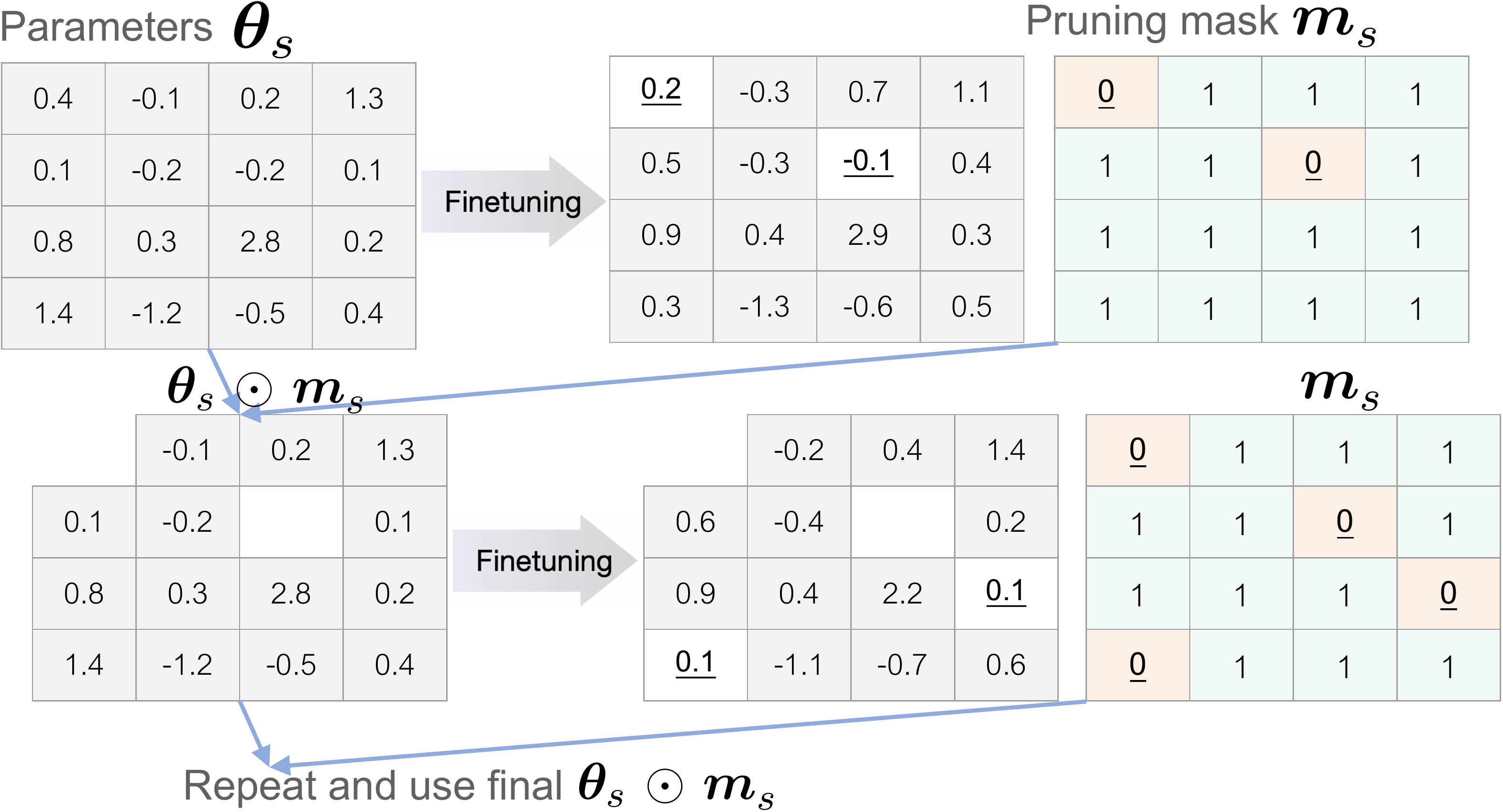}
   \caption{Overview of iterative magnitude pruning (Section~\ref{sec:imp}). We can also use regularizers during finetuning to diversify pruning (Section~\ref{sec:masked_reg}).}
   \label{fig:schemes_imp}
\end{figure}

We employ iterative magnitude pruning~\cite{frankle2018the} to find winning tickets for simplicity.
Other sophisticated options are left for future work.
Here, we explain the algorithm (refer to the paper for details).
The algorithm explores a good pruning mask via rehearsals of finetuning.
First, it completes a finetuning procedure of an initialized dense network and identifies the parameters with the 10\% lowest magnitudes as the targets of pruning.
Then, it makes the pruned subnetwork and resets its parameters to the originally-initialized (sub-)parameters.
This finetune-prune-reset process is repeated until reaching the desired pruning ratio.
We used 30\% as pruning ratio.

\subsection{Pruning with Regularizer}\label{sec:masked_reg}

We discussed that finetuning with different random seeds did not lead to diverse parameters in Section~\ref{sec:div_finetuning}.
Therefore, iterative magnitude pruning with different seeds could also produce less diverse subnetworks.
Thus, we also explore means of diversifying pruning patterns by enforcing different parameters to have lower magnitudes.
Motivated by this, we experiment with a simple approach, applying an $L_1$ regularizer (i.e., magnitude decay) to different parameters selectively depending on the random seeds.
Specifically, we explore two policies to determine which parameters are decayed and how strongly they are, i.e., the element-wise coefficients of the $L_1$ regularizer, $\bm{l}_s \in {\mathbb{R}_{\geq 0}}^{|\theta|}$.
During finetuning (for pruning), we add a regularization term $\tau ||\bm{\theta}_s \!\odot \bm{l}_s||_1$ with a positive scalar coefficient $\tau$ into the loss of the task (e.g., cross entropy for classification), where $\odot$ is element-wise product.
This softly enforces various parameters to have a lower magnitude among a set of random seeds and could lead various parameters to be pruned.

\paragraph{Active Masking}
To maximize the diversity of the surviving parameters of member models, it is necessary to prune the surviving parameters of the random seed $s_1$ when building a model with the next random seed $s_2$.
Thus, during finetuning with seed $s_2$, we apply the $L_1$ regularizer on the first surviving parameters.
Likewise, with the following seeds $s_3, s_4, ..., s_i, ..., s_{|\mathcal{M}|}$, we cumulatively use the average of the surviving masks as the regularizer coefficient mask.
Let $\bm{m}_{s_j} \in \{0, 1\}^{|\bm{\theta}|}$ be the pruning mask indicating surviving parameters from seed $s_j$, the coefficient mask with seed $s_i$ is $\bm{l}_{s_i} = \sum_{j < i} \bm{m}_{s_j} / (i-1)$.
We call this affirmative policy as \emph{active masking}.

\paragraph{Random Masking}
In active masking, each coefficient mask has a sequential dependence on the preceding random seeds.
Thus, the training of ensemble members cannot be parallelized.
Therefore, we also experiment with a simpler and parallelizable variant, \emph{random masking}, where a mask is independently and randomly generated from a random seed.
With a random seed $s_i$, we generate the seed-dependent random binary mask, i.e., $\bm{l}_s = \bm{m}^{\textrm{rand}}_{s_i} \in \{0, 1\}^{|\bm{\theta}|}$, where each element is sampled from Bernoulli distribution and 0's probability equals to the target pruning ratio.

\section{Experiments}

We evaluate the performance of ensembles using four finetuning schemes: (1) finetuning without pruning (\baseline{}), (2) finetuning of lottery-ticket subnetworks found with the naïve iterative magnitude pruning (\baselt{}), and (3) with $L_1$ regularizer by the active masking (\activelt{}) or (4) random masking (\randomlt{}).
We also compare with (5) \bagging{}-based ensemble, which trains dense models on different random 90\% training subsets.
We use the GLUE benchmark~\cite{wang2018glue} as tasks.
The implementation and settings follow \citet{chen2020bertlth}\footnote{
We found a bug in \citet{chen2020bertlth}'s implementation on GitHub, so we fixed it and experimented with the correct version.
} using the Transformers library~\cite{wolf2020transformers} and its bert-base-uncased pretrained model.
We report the average performance using twenty different random seeds.
Ensembles are evaluated using exhaustive combinations of five members.
We also perform Student's t-test for validating statistical significance\footnote{Note that not all evaluation samples satisfy independence assumption.}.
Note that, while the experiments focus on using BERT, we believe that the insights would be helpful to other pretrain-then-finetune settings in general\footnote{
\citet{raffel2020limittransfer} reported that the same problem happened on almost all tasks (GLUE~\cite{wang2018glue}, SuperGLUE~\cite{wang2019superglue}, SQuAD~\cite{rajpurkar2016squad}, summarization, and machine translation) using the T5 model.}.

\subsection{Accuracy}\label{sec:accuracy}

\setlength{\tabcolsep}{1.1mm}
\begin{table}[t]
\centering
\small
\begin{tabular}{l|lll|lll}
\hline
    & \multicolumn{3}{c|}{MRPC} & \multicolumn{3}{c}{STS-B} \\ 
        & single & ens. & diff. & single & ens. & diff. \\ \hline
\baseline{} & 83.48 & 84.34 & +0.86 & 88.35 & 89.04 & +0.69  \\
(\bagging{}) & 82.87 & 84.19 & +1.32 & 88.17 & 88.84 & +0.68  \\ \hline
\baselt{} & 83.84 & \bestorpar{\beatbase{84.98}} & \beatbase{+1.14} & 88.37 & \beatbase{89.16} & \beatbase{+0.79} \\ %
\activelt{} & 83.22 & \beatbase{84.60} & \beatblt{+1.38} & 88.39 & \bestorpar{\beatblt{89.32}} & \bestorpar{\beatblt{+0.94}} \\ %
\randomlt{} & 83.53 & \bestorpar{\beatbase{85.05}} & \bestorpar{\beatblt{+1.52}} & 88.49 & \bestorpar{\beatblt{89.35}} & \beatblt{+0.86} \\ \hline
\end{tabular}
\caption{The performances (single, ens.) and the improvements by ensembling (diff.).  \beatbase{Italic} indicates that the value is significantly larger than that of \baseline{}. \beatblt{Bold-italic} indicates significantly larger than that of both \baseline{} and \baselt{}.
\bestorpar{Underline} indicates the best.}
\label{tab:accuracy_table}
\end{table}

\begin{figure}[t]
  \centering 
  \includegraphics[width=\hsize]{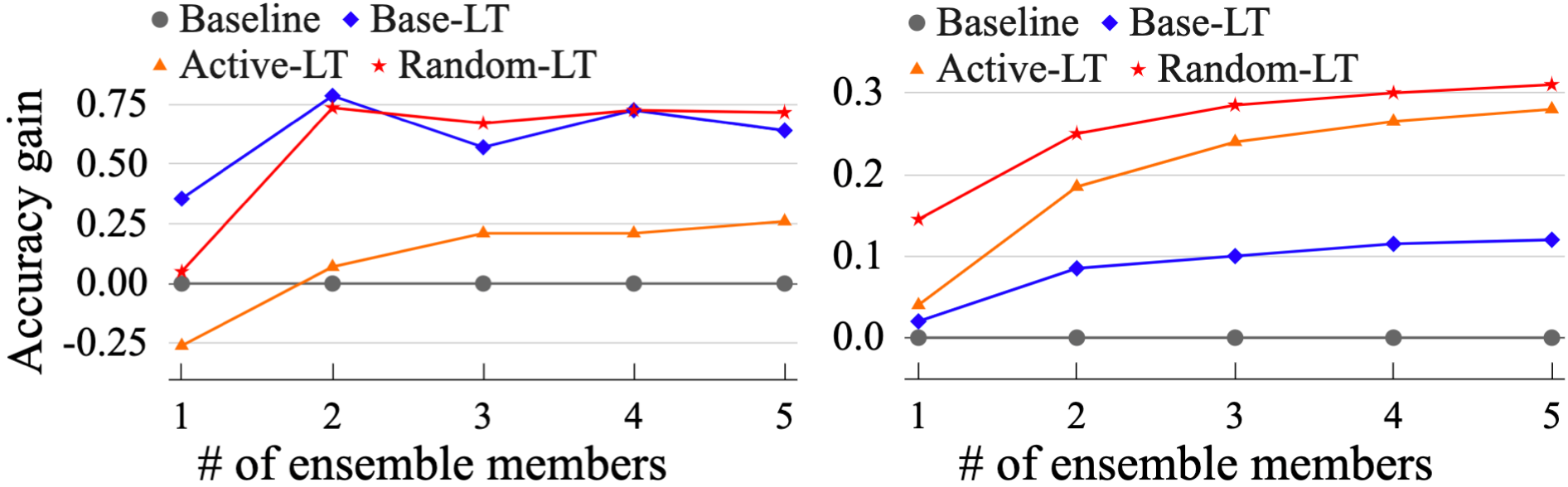}
   \caption{Comparison of the performances and the number of ensemble members on MRPC (left) and STS-B (right). They are represented as the relative gain compared with \baseline{}'s accuracy.}
   \label{fig:plot_e_num}
\end{figure}

We show the results on MRPC~\cite{dolan2005mrpc} and STS-B~\cite{cer2017stsb} in Table~\ref{tab:accuracy_table}.
Multi-ticket ensembles (\anylt{}) outperform \baseline{} and \bagging{} significantly ($p<0.001$).
This result supports the effectiveness of multi-ticket ensemble.
Note that the improvements of \anylt{} are attributable to ensembling (diff.) rather than to any performance gains of the individual models (single).
We also plot the improvements (ens. values relative to \baseline{}) as a function of the number of ensemble members on MRPC and STS-B in Figure~\ref{fig:plot_e_num}.
This also clearly shows that while the single models of \anylt{} have accuracy similar to \baseline{}, the gains appear when ensembling them.
While multi-ticket ensemble works well even with the naive pruning method (\baselt{}),
\randomlt{} and \activelt{} achieve the better ensembling effect on average; this suggests the effectiveness of regularizers.
Interestingly, \randomlt{} is simpler but more effective than \activelt{}.

\paragraph{When Winning Tickets are Less Accurate}
Does multi-ticket ensemble work well on any tasks?
The answer is no.
To enjoy the benefit from multi-ticket ensemble, we have to find diverse winning-ticket subnetworks sufficiently comparable to their dense network.
When winning tickets are less accurate than the baseline, their ensembles often fail to outperform the baseline's ensemble.
It happened to CoLA~\cite{warstadt2019cola}, QNLI~\cite{rajpurkar2016squad}, SST-2~\cite{socher2013sst2}, MNLI~\cite{williams2018mnli}; the naive iterative magnitude pruning did not find comparable winning-ticket subnetworks (with or sometimes even without regularizers)\footnote{
Although some studies~\cite{prasanna2020allwinningbert,chen2020bertlth,liang2021super} reported that they found winning-ticket subnetworks on these tasks, our finding did not contradict it.
Their subnetworks were often actually a little worse than their dense networks, as well as we found.
\citet{chen2020bertlth} defined winning tickets as subnetworks with performances within one standard deviation from the dense networks.
\citet{prasanna2020allwinningbert} considered subnetworks with even 90\% performance as winning tickets.
}\footnote{For example, comparing \baseline{} with \randomlt{} of pruning ratio 20\%, their average values of single/ensemble/difference are 91.38/91.93/+0.55 vs. 91.09/91.90/+0.81 on SST-2.
}\footnote{
This also happens to experiments with roberta-base while multi-ticket ensemble still works well on MRPC.
}.
Note that, even in such a case, \randomlt{} often yielded a higher effect of ensembling (diff.), while the degradation of single models canceled out the effect in total, and \bagging{} also failed to improve.
More sophisticated pruning methods~\cite{blalock2020prune,sanh2020movement} or tuning will find better winning-ticket subnetworks and maximize the opportunities for multi-ticket ensemble in future work.

\subsection{Diversity of Predictions}

As an auxiliary analysis of behaviors, we show that each subnetwork produces diverse predictions.
Because any existing diversity scores do not completely explain or justify the ensemble performance\footnote{Finding such a convenient diversity metric itself is still a challenge in the research community~\cite{wu2021revisitdiversity}.},
we discuss only rough trends in five popular metrics of classification diversity; Q statistic~\cite{yule1900qsta}, ratio errors~\cite{aksela2003comparison}, negative double fault~\cite{giacinto2001design}, disagreement measure~\cite{skalak96disagreement}, and correlation coefficient~\cite{kuncheva2003measures}.
See \citet{kuncheva2003measures,rafael2020deslib} for their summarized definitions.
As shown in Table~\ref{tab:diversity_table},
in all the metrics, winning-ticket subnetworks (\anylt{}) produced more diverse predictions than the baseline using the dense networks (\baseline{}).

\setlength{\tabcolsep}{1.6mm}
\begin{table}[t]
\centering
\small
\begin{tabular}{l|lllll} \hline
        & Q$\downarrow$ & R$\uparrow$ & ND$\uparrow$ & D$\uparrow$ & C$\downarrow$ \\ \hline
\baseline{} & 0.96 & 0.72 & -0.12 & 0.09 & 0.69 \\ \hline
\baselt{} & 0.93 & 1.00 & -0.11 & 0.10 & 0.62 \\ %
\activelt{} & 0.94 & 0.94 & -0.11 & 0.11 & 0.62 \\ %
\randomlt{} & 0.94 & 0.94 & -0.11 & 0.10 & 0.63 \\ \hline
\end{tabular}
\caption{Diversity metrics on MRPC. The signs, $\downarrow$ and $\uparrow$, indicate that the metric gets lower and higher when the predictions are diverse. Q = Q statistic, R = ratio errors, ND = negative double fault, D = disagreement measure, C = correlation coefficient.}
\label{tab:diversity_table}
\end{table}

\subsection{Diversity of Subnetwork Structures}

\begin{figure}[t]
  \centering 
  \includegraphics[width=\hsize]{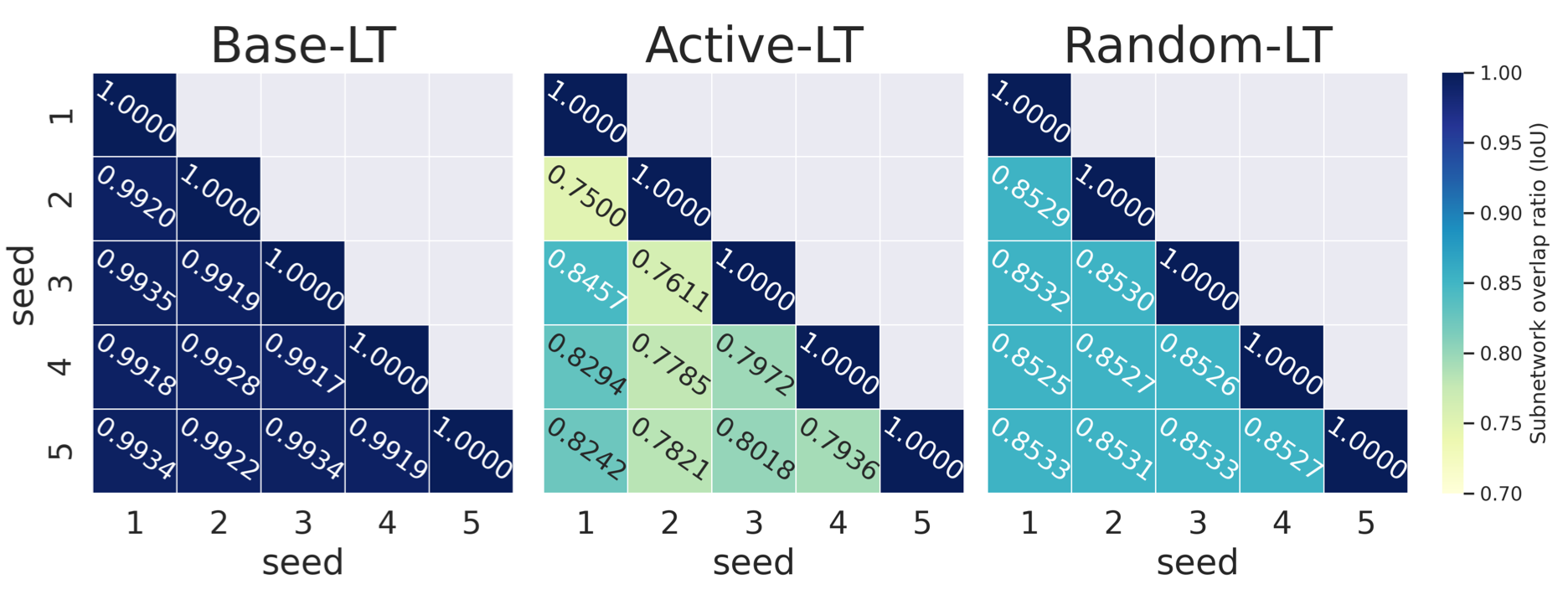}
   \caption{Overlap ratio of pruning masks $\bm{m}_{s_i}$ between different seeds on MRPC. The lower (yellower) the value is, the more dissimilar the two masks are.}
   \label{fig:overlapratio}
\end{figure}

We finally revealed the diversity of the subnetwork structures on MRPC.
We calculated the overlap ratio of two pruning masks, which is defined as intersection over union, $\textrm{IoU} = \frac{|\bm{m}_i \cap \bm{m}_j|}{|\bm{m}_i \cup \bm{m}_j|}$~\cite{chen2020bertlth}.
In Figure~\ref{fig:overlapratio}, we show the overlap ratio between the pruning masks for the five random seeds, i.e., $\{\bm{m}_{s_1}, ..., \bm{m}_{s_{5}}\}$. %
At first, we can see that \activelt{} and \randomlt{} using the regularizers resulted in diverse pruning. %
This higher diversity could lead to the best improvements by ensembling, as discussed in Section~\ref{sec:accuracy}.
Secondly, \baselt{} produced surprisingly similar (99\%) pruning masks with different random seeds.
However, recall that even \baselt{} using the naïve iterative magnitude pruning performed better than \baseline{}.
This result shows that even seemingly small changes in structure can improve the diversity of predictions and the performance of the ensemble.

\section{Related Work}\label{app:relatedwork}

Some concurrent studies also investigate the usage of subnetworks for ensembles.
\citet{gal16mcdropout} is a pioneer to use subnetwork ensemble.
A trained neural network with dropout can infer with many different subnetworks, and their ensemble can be used for uncertainty estimation, which is called MC-dropout.
\citet{durasov2021masksembles} improved the efficiency of MC-dropout by exploring subnetworks.
\citet{zhang2021ex} (unpublished) experimented with an ensemble of subnetworks of different structures and initialization when trained from scratch, while the improvements possibly could be due to regularization of each single model.
\citet{havasi2021training} is a similar but more elegant approach, which does not explicitly identify subnetworks.
Instead, it trains a single dense model with training using multi-input multi-output inference; the optimization can implicitly find multiple disentangled subnetworks in the dense model during optimization from random initialization.
These studies support our assumption that different subnetworks can improve ensemble by diversity.
\citet{liu2022deepnoverhead} efficiently trains multiple subnetworks, whose ensemble is competitive with dense ensembles.

Some other directions for introducing diversity exist, while most are unstable.
Promising directions are to use entropy~\cite{pang19impadv} or adversarial training~\cite{rame2021dice}.
Although they required complex optimization processes, they improved the robustness or ensemble performance on small image recognition datasets.

Recently, concurrent work~\cite{sellam2022multibert,tay2022scalet5} provide multiple BERT or T5 models pretrained from different seeds or configurations for investigation of seed or configuration dependency using large-scale computational resources.
Further research with the models and such computational resources will be helpful for more solid comparison and analysis.

Note that no prior work tackled the problem of ensembles from a pre-trained model.
Framing the problem is one of the contributions of this paper.
Secondly, our multi-ticket ensemble based on random masking enables an independently parallelizable training while existing methods require a sequential processing or a grouped training procedure.
Finally, multi-ticket ensemble can be combined with other methods, which can improve the total performance together.

\section{Conclusion}

We raised a question on difficulty of ensembling large-scale pretrained models.
As an efficient remedy, we explored methods to use subnetworks in a single model.
We empirically demonstrated that ensembling winning-ticket subnetworks could outperform the dense ensembles via diversification and indicated a limitation too.

\section*{Acknowledgments}

We appreciate the helpful comments from the anonymous reviewers.
This work was supported by JSPS KAKENHI Grant Number JP19H04162.

\bibliography{anthology,custom}
\bibliographystyle{acl_natbib}

\clearpage
\appendix

\section{The Setting of Fine-tuning}\label{app:settings}
We follow the setting of \citet{chen2020bertlth}'s implementation; epoch: 3, initial learning rate: 2e-5 with linear decay, maximum sequence length: 128, batch size: 32, dropout probability: 0.1.
This is one of the most-used settings for finetuning a BERT; e.g., the example of finetuning in the Transformers library~\cite{wolf2020transformers} uses the setting\footnote{\url{https://github.com/huggingface/transformers/blob/7e406f4a65727baf8e22ae922f410224cde99ed6/examples/pytorch/text-classification/README.md\#glue-tasks}}.

We did not prune the embedding layer, following \citet{chen2020bertlth,prasanna2020allwinningbert}.
The coefficient of $L_1$ regularizer, $\tau$, is decayed using the same scheduler as the learning rate.
We tuned it on MRPC and used it for other tasks.

\section{The Learning Rate Scheduler of \citet{chen2020bertlth}}
\label{app:chenbug}
Our implementation used in the experiments are derived from \citet{chen2020bertlth}'s implementation\footnote{\url{https://github.com/VITA-Group/BERT-Tickets}}.
However, we found a bug in \citet{chen2020bertlth}'s implementation on GitHub.
Thus, we fixed it and experimented with the correct version.
In their implementation, the learning rate schedule did not follow the common setting and the description mentioned in the paper; \textit{`We use standard implementations and hyperparameters [49]. Learning rate decays linearly from initial value to zero'}.
Specifically, the learning rate with linear decay did not reach zero but was at significant levels even at the end of the finetuning.
Our implementation corrected it so that it did reach zero as specified in their paper and in the common setting.

\section{The Combinations of Ensembles}
\label{app:combinations_of_ensembles}
In the experiments, we first prepared twenty random seeds and split them into two groups, each of which trained ten models.
For stabilizing the measurement of the result, we exhaustively evaluated all the possible combinations of ensembles (i.e., depending on the number of members, ${}_{10} \mathrm{C} _2$, ${}_{10} \mathrm{C} _3$, ${}_{10} \mathrm{C} _4$, ${}_{10} \mathrm{C} _5$ patterns, respectively) among the ten models for each group, and averaged the results with the two groups.
The performance of the members is also averaged over all the seeds.

\section{The Results with RoBERTa}\label{app:roberta}

We simply conducted supplementary experiments with RoBERTa~\cite{liu2019roberta} (robeta-base model), although optimal hyperparameters were not searched well.
The results were similar to the cases of base-base-uncased.
The patterns can be categorized into the three.
First, multi-ticket ensembles worked well with roberta on MRPC, as shown in Table~\ref{tab:accuracy_table_roberta}.
Secondly, accurate winning-ticket subnetworks were not found on CoLA and QNLI.
Although the effect of ensembleing was improved after pruning, each single model got worse and the final ensemble accuracy did not outperform the dense baseline.
Thirdly, although accurate winning-ticket subnetworks were found on STS-B and SST-2, regularizations worsened single-model performances.
While this case also improved the effect of ensembling, the final accuracy did not outperform the baseline.
These experiments further emphasized the importance of development of more sophisticated pruning methods without sacrifice of model performances in the context of the lottery ticket hypothesis.

\setlength{\tabcolsep}{1.1mm}
\begin{table}[t]
\centering
\small
\begin{tabular}{l|lll|lll}
\hline
    & \multicolumn{3}{c|}{MRPC} & \multicolumn{3}{c}{STS-B} \\ 
        & single & ens. & diff. & single & ens. & diff. \\ \hline
\baseline{} & 87.77 & 88.47 & +0.70 & 89.52 & 90.00 & +0.48  \\
(\bagging{}) & 87.64 & 88.12 & +0.49 & 89.34 & 89.91 & +0.54  \\ \hline
\baselt{} & 87.72 & 88.25 & +0.53 & 89.71 & 90.07 & +0.36 \\ %
\activelt{} & 87.39 & 88.51 & +1.12 & 88.46 & 89.50 & +1.04 \\ %
\randomlt{} & 87.86 & 89.26 & +1.40 & 88.41 & 89.39 & +0.98 \\ \hline
\end{tabular}
\caption{The performances (single, ens.) and the improvements by ensembling (diff.) of RoBERTa-base models.}
\label{tab:accuracy_table_roberta}
\end{table}

\end{document}